\documentclass[a4paper, 10pt, conference]{ieeeconf}      

\IEEEoverridecommandlockouts                              

\usepackage{graphics} 
\usepackage{amsmath} 
\usepackage{amssymb}  
\usepackage{amsthm}
\usepackage{psfrag}
\usepackage{graphicx}
\usepackage{float}

\newcommand{\comment}[1]{}

\theoremstyle{plain}

\theoremstyle{definition}

\newtheorem{thm}{\protect\theoremname}
\newtheorem{defn}[thm]{\protect\definitionname}
\providecommand{\definitionname}{Definition}
\providecommand{\theoremname}{Theorem}

\newcommand\mycomment[1]{} 			     

\newcommand{\ignore}[1]{}

\title{\LARGE \bf Deep Learning and the Information Bottleneck Principle}

\author{ Naftali Tishby$^{1,2}$ \and Noga Zaslavsky$^1$ 
\thanks{$^1$The Edmond and Lilly Safra Center for Brain Sciences, Hebrew University of Jerusalem, Israel.}%
\thanks{$^2$ School of Engineering and Computer Science, Hebrew University of Jerusalem, Israel.
{\tt\small tishby@cs.huji.ac.il}}%
\thanks{* This study was funded by the
Israeli Science Foundation center of excellence, the DARPA MSEE project, 
the Israeli Ministry of Science and Technology, and the
Intel Collaborative Research Institute for Computational Intelligence (ICRI-CI).}
}

\begin{document}

\maketitle
\thispagestyle{empty}
\pagestyle{empty}

\begin{abstract}
Deep Neural Networks (DNNs) are analyzed via the theoretical framework of the information bottleneck (IB) principle. We first show that any DNN can be quantified by the mutual information between the layers and the input and output variables.  Using this representation we can calculate the optimal information theoretic limits of the DNN and obtain finite sample generalization bounds. The advantage of getting closer to the theoretical limit is quantifiable both by the generalization bound and by the network's simplicity. We argue that both the optimal architecture, number of layers and features/connections at each layer, are related to the bifurcation points of the information bottleneck tradeoff, namely, relevant compression of the input layer with respect to the output layer. The hierarchical representations at the layered network naturally correspond to the structural phase transitions along the information curve. We believe that this new insight can lead to new optimality bounds and deep learning algorithms.
\end{abstract}



\section{Introduction}

Deep Neural Networks (DNNs) and Deep Learning (DL) algorithms in various
forms have become the most successful machine learning method for
most supervised learning tasks. Their performance currently surpass
most competitor algorithms and DL wins top machine learning competitions
on real data challenges \cite{Bengio2013a,HG-Sr-2006,krizhevsky2012imagenet}. 
The theoretical understanding of DL remains, however, unsatisfactory. Basic questions
about the design principles of deep networks, the optimal architecture,
the number of required layers, the sample complexity, and the best
optimization algorithms, are not well understood.

One step in that direction was recently made in a remarkable paper
by Metha and Schwab \cite{MP-SD-2014} that showed an exact mapping
between the variational Renormalization Group (RG) and DNNs based on 
Restricted Boltzmann Machines (RBMs). An important insight provided by that paper is that features along
the layers become more and more statistically decoupled as the layers
gets closer to the RG fixed point.

In this work we express this important insight using information
theoretic concepts and formulate the goal of deep learning as an information
theoretic tradeoff between compression and prediction. We first argue that the goal of 
any supervised learning is to capture and efficiently represent the relevant information
in the input variable about the output - label - variable. 
Namely, to extract an approximate 
minimal sufficient statistics of the input with respect to the output. The information 
theoretic interpretation of minimal sufficient statistics \cite{Cover1991} suggests a principled 
way of doing that: find a  maximally compressed mapping of the input variable that preserves 
as much as possible the information on the output variable. This is precisely the goal of the 
Information Bottleneck (IB) method \cite{NT-FP-WB-1999}.

Several interesting issues arise when applying this principle to DNNs. First, the layered
structure of the network generates a successive Markov chain of intermediate representations, 
which together form the (approximate) sufficient statistics. This is closely related to  
successive refinement of information in Rate Distortion Theory \cite{WE-TC-1991}.
Each layer in the network can now be quantified by the amount of information it retains on the 
input variable, on the (desired) output variable, as well as on the predicted output of the network.   
The Markovian structure and data processing inequalities enable us to examine the efficiency 
of the internal representations of the network's hidden layers, which is not possible with other 
distortion/error measures. It also provides us with the information theoretic limits of the 
compression/prediction problem and theoretically quantify each proposed DNN for the given training 
data. In addition, this representation of DNNs gives a new theoretical sample complexity bound,
using the known finite sample bounds on the IB \cite{OS-SS-NT-2010}.
    
Another outcome of this representation is a possible explanation of the layered 
architecture of the network, different from the one suggested in \cite{MP-SD-2014}.  
Neurons, as non-linear (e.g. sigmoidal) functions of a dot-product of their input, can only capture
linearly separable properties of their input layer.  Linear separability is  possible when the
input layer units are close to conditional independence, given the output classification. 
This is generally not true for the data distribution and intermediate hidden layer are required.
We suggest here that the break down of the linear-separability is associated with a representational 
phase transition (bifurcation) in the IB optimal curve, as both result from the 
second order dependencies in the data. Our analysis suggests new information theoretic optimality conditions, sample complexity bounds, and design principle for DNN models. 

The rest of the paper is organized as follows. We first review the structure of DNNs as a Markov cascade
of intermediate representations between the input and output layers, made out of layered sigmoidal neurons.
Next we review the IB principle as a special type of Rate Distortion problem, and 
discuss how DNNs can be analyzed in this special rate-distortion distortion plane. In section III we describe
the information theoretic constraints on DNNs and suggest a new optimal learning principle, using finite sample 
bounds on the IB problem. Finally, we suggest an intriguing connection between the IB structural phase transitions 
and the layered structure of DNNs.

\ignore{The role of the
other layers in deep learning is thus to remove the redundancies in
the input layers, namely, to compress them, without loss of information
relevant to the classification task. We argue that the latter is optimally
captured by constraining the mutual information between the hidden
layers representation and the output classification.}

\section{Background}

\subsection{Deep Neural Networks}

\begin{figure}[t]
\begin{centering}
\includegraphics[scale = 0.3]{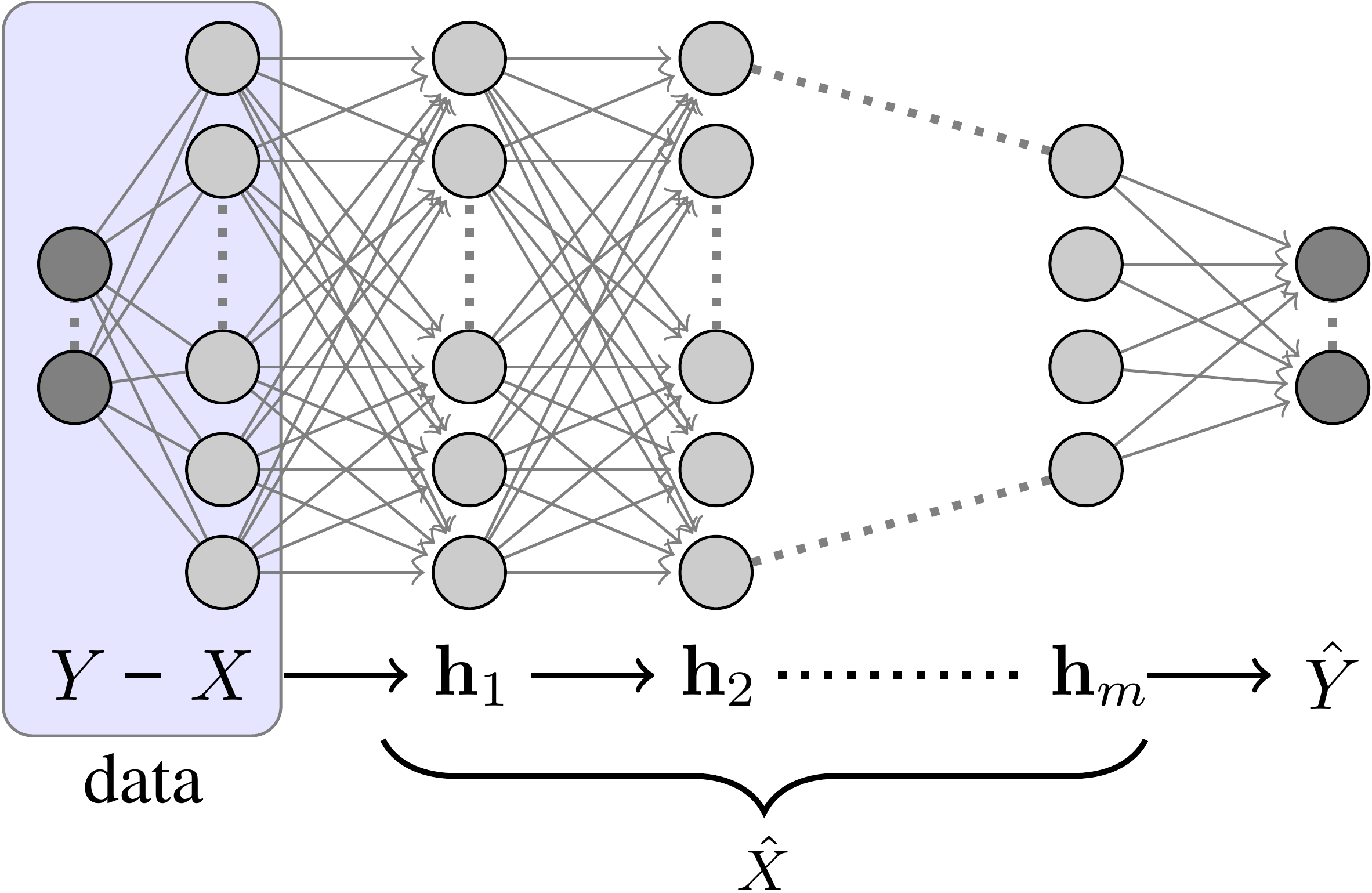}
\par\end{centering}
\caption{An example of a feedforward DNN with $m$ hidden layers, an input layer $X$ and an output layer $\hat{Y}$. The desired output, $Y$, is observed only during the training phase through a finite sample of the joint distribution, $p(X,Y)$, and is used for learning the connectivity matrices between consecutive layers. After training, the network receives an input $X$, and successively processes it through the layers, which form a Markov chain, to the predicted output $\hat{Y}$. $I(Y;\hat{Y})/I(X;Y)$ quantifies how much of the relevant information is captured by the network.}
\label{fig-DNN}
\end{figure}

DNNs are comprised of multiple layers of artificial neurons, or simply
units, and are known for their remarkable performance in learning
useful hierarchical representations of the data for various machine
learning tasks. While there are many different variants of DNNs \cite{YB-2009}, here
we consider the rather general supervised learning settings of feedforward networks 
in which multiple hidden layers separate the input and output layers of the network
(see figure \ref{fig-DNN}). Typically, the input, denoted by $X$, is a high dimensional
variable, being a low level representation of the data such as pixels of an image, whereas the
desired output, $Y$, has a significantly lower dimensionality of the predicted categories. 
This generally means that most of the entropy of $X$ is not very informative about
$Y$, and that the relevant features in $X$ are highly distributed
and difficult to extract. The remarkable success of DNNs in learning to extract
such features is mainly attributed to the sequential processing of the data,
namely that each hidden layer operates as the input to the next one,
which allows the construction of higher level distributed representations.

The computational ability of a single unit in the network is limited,
and is often modeled as a sigmoidal neuron. This means that the output
of each layer is $\mathbf{h}_{k}=\sigma\left(W_{k}{\bf h}_{k-1}+{\bf b}_{k}\right)$,
where $W_{k}$ is the connectivity matrix which determines the weights
of the inputs to ${\bf h}_{k}$, ${\bf b}_{k}$ is a bias term, and
$\sigma(u)=\frac{1}{1+\exp(-u)}$ is the standard sigmoid function. 
Given a particular architecture, training the network is reduced to learning the weights between each
layer. This is usually done by stochastic gradient decent methods, such as back-propagation,
that aim at minimizing some prediction error, or distortion, between the desired and 
predicted outputs $Y$ and $\hat{Y}$ given the input $X$. 
Interestingly, other DNN architectures implement stochastic mapping between the layers, such as 
the RBM based DNNs \cite{HG-Sr-2006}, but it is not clear so far why or when such 
stochasticity can improve performance. 
Symmetries of the data are often taken into account through weight sharing, as in convolutional neural networks \cite{lecun1995convolutional,krizhevsky2012imagenet}. 

Single neurons can (usually) classify only linearly separable inputs,
as they can implement only hyperplanes in their input space, ${\bf u}={\bf w}\cdot {\bf h}+{\bf b}$. 
Hyperplanes can optimally classify data when the inputs are conditionally independent. 
To see this, let $p({\bf x}|y)$ denote the (binary) class ($y$) conditional probability of the inputs ${\bf x}$. 
Bayes theorem tells us that
\begin{equation}
p(y|{\bf x})=\frac{1}{1+\exp\left(-\log{\frac{p({\bf x}| y)}{p({\bf x}| y')}}-\log{\frac{p(y)}{p(y')}}\right)}
\end{equation}
which can be written as a sigmoid of a dot-product of the inputs when 
\begin{equation}
\frac{p({\bf x} | y)}{p({\bf x} |y') }=\prod_{j=1}^N \left[\frac{p(x_j|y)}{p(x_j|y')}\right]^{np(x_j)}
\label{eq:Lin-Sep} ~.
\end{equation}
The sigmoidal neuron can calculate precisely the posterior probability with weights 
$w_j=\log{\frac{p(x_j|y)}{p(x_j|y')}}$, and bias $b=\log{\frac{p(y)}{p(y')}}$, when the neuron's inputs are 
proportional to the probability of the respective feature in the input layer, i.e.  $h_j = n p(x_j)$. 
As such conditional independence can not be assumed for general data distributions, 
representational changes through the hidden layers are required, up to linear transformation 
that can decouple the inputs. 

As suggested in \cite{MP-SD-2014}, approximate conditional independence is effectively 
achieved for RBM based DNNs through successive RG transformations that decouple the units without loss of
relevant information. The relevant compression, however, is implicit in the RG transformation and does not 
hold for more general DNN architectures.

The other common way of statistically decoupling the units is by dimension expansion, 
or embedding in very high dimension, as done implicitly by Kernel machines, or by random expansion. 
There are nevertheless sample and computational costs to such dimensional expansion 
and these are clearly not DNN architectures.    

In this paper we propose a purely information theoretic view of DNNs, which can quantify their 
performance, provide a theoretical limit on their efficiency, 
and give new finite sample complexity bounds on their generalization abilities. Moreover, our analysis
suggests that the optimal DNN architecture is also determined solely by an information theoretic analysis 
of the joint distribution of the data, $p(X,Y)$.

\ignore{Here we propose using an explicit information theoretic compression of the relevant information, using the 
Information Bottleneck principle for learning the complete architecture and representations of the 
network in a principled way. It turns out that there is a direct connection between the linear separability
requirement and the relevant compression of the IB, which suggests that the DNN hidden layers correspond to 
the structural phase transitions obtained during the IB compression.}

\subsection{The Information Bottleneck Principle}

The information bottleneck (IB) method was introduced 
as an information theoretic principle for extracting relevant information
that an input random variable $X\in{\cal X}$ contains about an output
random variable $Y\in{\cal Y}$. Given their joint distribution $p\left(X,Y\right)$,
the \textit{relevant information} is defined as the mutual information
$I\left(X;Y\right)$, where we assume statistical dependence between $X$ and $Y$. 
In this case, $Y$ implicitly determines the relevant and irrelevant features in $X$. 
An optimal representation of $X$ would capture the relevant features, and compress $X$ by
dismissing the irrelevant parts which do not contribute to the prediction of $Y$.

In pure statistical terms, the relevant part of $X$ with respect to $Y$, denoted by $\hat{X}$, is 
a {\em minimal sufficient statistics} of $X$ with respect $Y$. Namely, it is the simplest mapping 
of $X$ that captures the mutual information $I(X;Y)$. 
We thus assume the Markov chain $Y\rightarrow X\rightarrow\hat{X}$ and minimize the mutual 
information $I(X;\hat{X})$ to obtain the simplest statistics (due to the data processing inequality (DPI) \cite{Cover1991}), 
under a constraint on $I(\hat{X};Y)$. Namely, finding an optimal representation 
$\hat{X}\in{\cal \hat{X}}$ is formulated as the minimization of the following Lagrangian 
\begin{equation}
{\cal L}\left[p\left(\hat{x}|x\right)\right]=I\left(X;\hat{X}\right)-\beta I\left(\hat{X};Y\right)\label{eq:IB-1}
\end{equation}
subject to the Markov chain constraint. The positive Lagrange multiplier $\beta$ 
operates as a tradeoff parameter between the complexity (rate) of the representation,
$R=I(X;\hat{X})$, and the amount of preserved relevant information,
$I_{Y}=I(\hat{X};Y)$. 

For general distributions, $p(X,Y)$, exact minimal sufficient statistics may not exist, and the prediction
Markov chain, $X\rightarrow\hat{X}\rightarrow Y$ is incorrect. If we denote by $\hat{Y}$ the predicted 
variable, the DPI implies $I(X;Y)\ge I(Y;\hat{Y})$, with equality if and only if $\hat{X}$ is a sufficient statistic.

As was shown in \cite{NT-FP-WB-1999}, the optimal solutions for the IB variational problem 
satisfy the following self-consistent equations for some value of $\beta$,
\begin{eqnarray}
p\left(\hat{x}|x\right) & = & \frac{p\left(\hat{x}\right)}{Z\left(x;\beta\right)}\exp\left(-\beta D\left[p\left(y|x\right)\|p\left(y|\hat{x}\right)\right]\right)\nonumber\\
p\left(y|\hat{x}\right) & = & \sum_{x}p\left(y|x\right)p\left(x|\hat{x}\right)\label{IB_cent}\nonumber\\
p\left(\hat{x}\right) & = & \sum_{x}p\left(x\right)p\left(\hat{x}|x\right)\nonumber
\end{eqnarray}
where $Z\left(x;\beta\right)$ is the normalization factor, also known as the partition function.

The IB can be seen as a rate-distortion problem with a non-fixed distortion measure that depends on the
optimal map, defined as 
$d_{IB}\left(x,\hat{x}\right)=D\left[p\left(y|x\right)\|p\left(y|\hat{x}\right)\right]$,
where $D$ is the Kullback-Leibler divergence. The self consistent equations can be iterated, as in the Arimoto-Blahut
algorithm, for calculating the optimal IB tradeoff, or rate-distortion function, though this is not a convex
optimization problem.

With this interpretation, the expected IB distortion is then 
\[
D_{IB}=E\left[d_{IB}\left(X,\hat{X}\right)\right]=I(X;Y|\hat{X})
\]
which is the residual information between $X$ and $Y$, namely the
relevant information  {\em not} captured by $\hat{X}$.  
Clearly, the variational principle in Eq.\ref{eq:IB-1} is equivalent to
\[
\tilde{{\cal L}}\left[p\left(\hat{x}|x\right)\right]=I\left(X;\hat{X}\right)+\beta I\left(X;Y\right|\hat{X})
\]
as they only differ by a constant. The optimal tradeoff for this variational problem
is defined by a rate-distortion like curve \cite{Gilad-Bachrach03aninformation},
as depicted by the black curve in figure \ref{fig-info-curve}. The parameter $\beta$ is the negative inverse 
slope of this curve, as with rate-distortion functions.

\ignore{In addition, if the cardinality of $\hat{X}$ is smaller than the cardinality of $X$ [TODO: should it be only less than $|X|-2$?], then the best achievable tradeoff would be optimal only for sufficiently small $\beta$, which can be seen by the sub-optimal curves in figure \ref{fig-info-curve}.}

Interestingly, the IB distortion curve, also known as the information curve for the joint distribution $p(X,Y)$,
may have bifurcation points to sub-optimal curves (the short blue curves in figure \ref{fig-info-curve}), 
at critical values of $\beta$. These bifurcations correspond to phase
transitions between different topological representations of $\hat{X}$, such as different cardinality in clustering 
by deterministic annealing \cite{KR-98}, or dimensionality change for continues variables \cite{GC-AG-NT-YW-2005}.  
These bifurcations are pure properties of the joint distribution, independent of any modeling assumptions.

Optimally, DNNs should learn to extract the most efficient informative features, or approximate 
minimal sufficient statistics, with the most compact architecture (i.e. minimal number of layers, 
with minimal number of units within each layer).

\section{A new Information Theoretic Learning Principle for DNNs}

\subsection{Information characteristics of the layers}

As depicted in figure \ref{fig-DNN}, each layer in a DNN processes inputs
only from the previous layer, which means that the network layers form a Markov chain. 
An immediate consequence of  the DPI  is
that information about $Y$ that is lost in one layer cannot be recovered
in higher layers. Namely, for any $i\ge j$ it holds that 
\begin{equation}
I\left(Y;X\right)\ge I\left(Y;\mathbf{h}_{j}\right)\ge I\left(Y;\mathbf{h}_{i}\right)\ge I\left(Y;\hat{Y}\right)
\label{eq:DNN-DPI}~.
\end{equation}
Achieving equality
in Eq.\ref{eq:DNN-DPI} is possible if and only if each layer is a
sufficient statistic of its input.  By requiring not only the most relevant
representation at each layer, but also the most concise representation
of the input, each layer should attempt to maximize $I\left(Y;\mathbf{h}_{i}\right)$
while minimizing $I\left(\mathbf{h}_{i-1};\mathbf{h}_{i}\right)$
as much as possible.

From a learning theoretic perspective, it may not be immediately clear
why the quantities $I\left(\mathbf{h}_{i-1};\mathbf{h}_{i}\right)$
and $I\left(Y;\mathbf{h}_{i}\right)$ are relevant for efficient learning
and generalization. 
\ignore{In the case of binary classification, it is well known
that $I\left(Y;\mathbf{h}_{i}\right)$ is related to an upper bound
on the Bayesian prediction error (known as the Raviv-Hellman inequality
\mycomment{TODO - cite?}, namely the maximum a-posteriori error when trying
to predict $Y$ from $\mathbf{h}_{i}$. In addition,}
It has been shown in \cite{OS-SS-NT-2010} that the mutual information $I(\hat{X};Y)$,
which corresponds to $I\left(Y;\mathbf{h}_{i}\right)$ in our context,
can bound the prediction error in classification tasks with multiple classes.  
In sequential multiple hypotheses testing, the mutual information gives a (tight) bound  
on the harmonic mean of the $\log$ probability of error over the decision time.

Here we consider $I(Y;\hat{Y})$ as the natural quantifier of the quality of the DNN, as 
it measures precisely how much of the predictive features in $X$ for $Y$ is 
captured by the model. 
Reducing $I\left(\mathbf{h}_{i-1};\mathbf{h}_{i}\right)$
also has a clear learning theoretic interpretation as the minimal description
length of the layer. 

The information distortion of the IB principle 
provides a new measure of optimality
which can be applied not only for the output layer, as done
when evaluating the performance of DNNs with other distortion or error measures, 
but also for evaluating the optimality of each hidden layer or unit of the network.
Namely, each layer can be compared to the optimal
IB limit for some $\beta$, 
\[
I\left(\mathbf{h}_{i-1};\mathbf{h}_{i}\right)+\beta I\left(Y;\mathbf{h}_{i-1}|\mathbf{h}_{i}\right)
\]
where we define $\mathbf{h}_{0}=X$ and $\mathbf{h}_{m+1}=\hat{Y}$.
This optimality criterion also give a nice interpretation of the construction
of higher level representations along the network. Since each point
on the information curve is uniquely defined by $\beta$, shifting
from low  to higher level representations is
analogous to successively decreasing $\beta$.  Notice that other
cost functions, such as the squared error, are not applicable for
evaluating the optimality of the hidden layers, nor can they account
for multiple levels of description.

\begin{figure}[t]
\begin{centering}
\includegraphics[trim = 2.9cm 1.5cm 2cm 1cm, scale = 0.37]{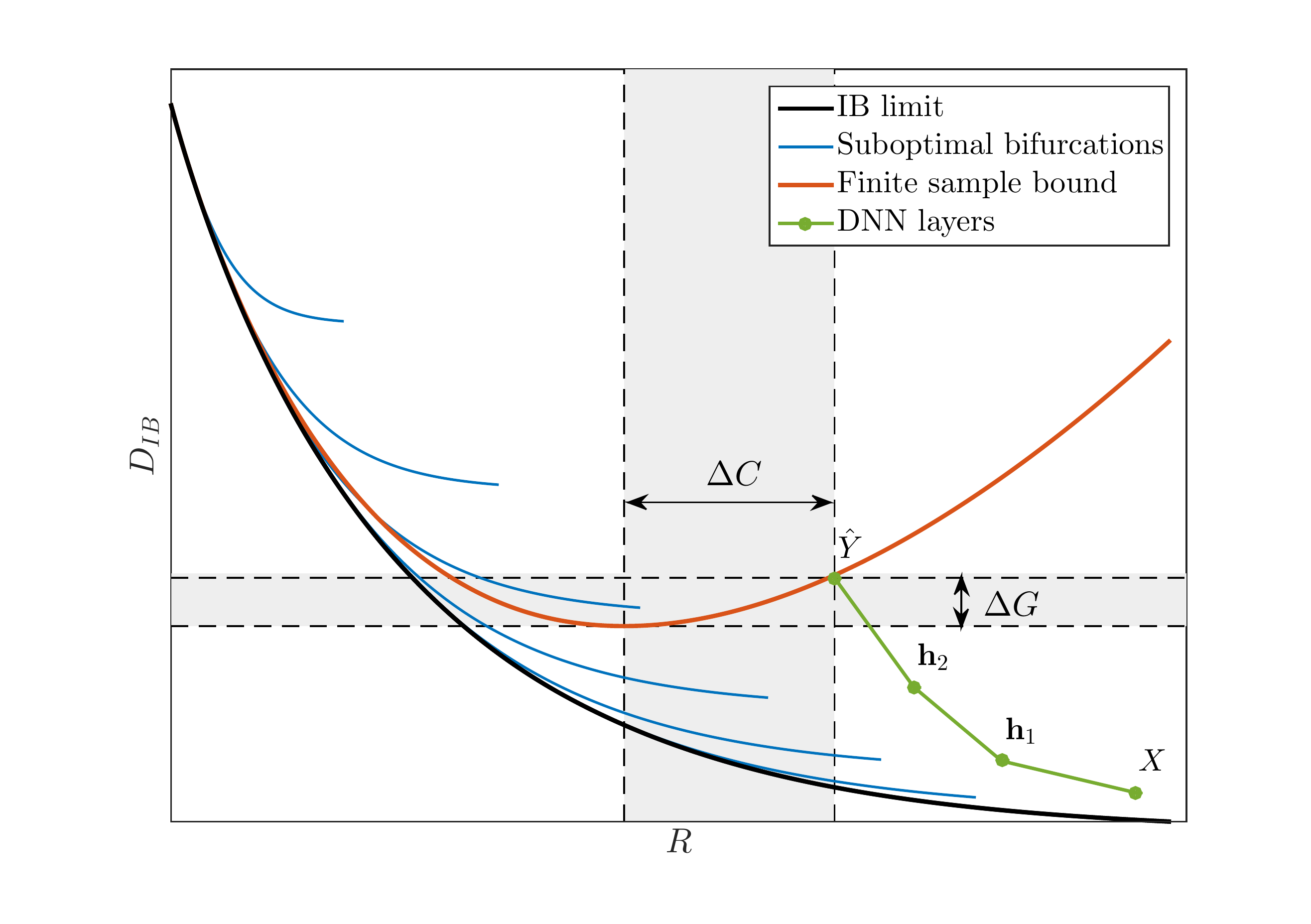}
\par\end{centering}
\caption{A qualitative information plane, with a hypothesized path of the layers in a typical DNN (green line) on the training data. The black line is the optimal achievable IB limit, and the blue lines are sub-optimal IB bifurcations, obtained by forcing the cardinality of $\hat{X}$ or remaining in the same representation. The red line corresponds to the upper bound on the {\em out-of-sample} IB distortion (mutual information on $Y$), when training from a finite sample. While the training distortion may be very low (the green points) the actual distortion can be as high as the red bound. This is the reason why
one would like to shift the green DNN layers closer to the optimal curve to obtain lower complexity and better generalization. Another interesting consequence is that getting closer to the optimal limit requires stochastic mapping between the layers. }
\label{fig-info-curve}
\end{figure}

The theoretical IB limit and the limitations that are imposed by the DPI on the flow of information between the layers, 
gives a general picture as to to where each layer of a trained network can be on the information plane. 
The input level clearly has the
least IB distortion, and requires the longest description (even after dimensionality reduction, $X$ is the lowest representation level in the network). Each consecutive layer can only increase the
IB distortion level, but it also compresses its inputs, hopefully  eliminating only irrelevant information. 
The green line in figure \ref{fig-info-curve} shows
a possible path of the layers in the information plane.

\subsection{Finite Samples and Generalization Bounds}

It is important to note that the IB curve is a property of the joint
distribution $p\left(X,Y\right)$, however this distribution is obviously unknown
in actual machine learning tasks. In fact, machine learning algorithms,
and in particular training algorithms for DNNs, have only access to
a finite sample. Nonetheless, it has been shown in \cite{OS-SS-NT-2010}
that it is possible to generalize using the IB principle as a learning
objective from finite samples, as long as the representational complexity
(i.e. the cardinality of $\hat{X}$) is limited. Assume all variables
have finite support, and let $K=|\hat{{\cal X}}|$. Denote by $\hat{I}$
the empirical estimate of the mutual information based on the finite sample
distribution $\hat{p}(x,y)$ for a given sample of size $n$. The
generalization bounds proven in \cite{OS-SS-NT-2010} guarantee that
\[
I\left(\hat{X};Y\right)\le\hat{I}\left(\hat{X};Y\right)+O\left(\frac{K\left|{\cal Y}\right|}{\sqrt{n}}\right)
\]
and that
\[
I\left(X;\hat{X}\right)\le\hat{I}\left(X;\hat{X}\right)+O\left(\frac{K}{\sqrt{n}}\right) ~.
\]

Notice that these bounds get worse with $K$, but do not depend on the cardinality of $X$. 
This means that the IB optimal curve can be well estimated for learning
compressed representations, and is badly estimated for learning complex models.
The complexity of the representation is not precisely the cardinality imposed by the support of $\hat{X}$, 
but its effective description length, namely $K \approx 2^{I\left(\hat{X};X\right)}$.
This gives a continuous worst case upper bound on the 
true $I(\hat{X};Y)$ for any given sample size $n$. 
This bound is illustrated in figure \ref{fig-info-curve}, when
interpreting the information curve (in black) as the empirical curve (i.e.
the optimal tradeoff with respect to $\hat{p}\left(X,Y\right)$ rather
than $p\left(X,Y\right)$). The red curve is the worst-case bound,
and its minimum is the optimal point on the information curve in the
sense that it gives the best worst case {\em true} tradeoff between the
complexity and the accuracy of the representation. Denote this point
by $\left(R^{*}\left(n\right),D_{IB}^{*}\left(n\right)\right)$. Notice
that the empirical information curve might be too optimistic 
especially at its extreme - most complex - end. Thus that point
is not truly the most informative, as opposed to corresponding point
on the true information curve.

From this analysis it is clear that the empirical input layer of a DNN alone
cannot guarantee good generalization even though it contains
more information about the target variable $Y$ than the hidden layers, as its representation of the data
is too complex. Compression
is thus necessary for generalization. In other words, the hidden layers
must compress the input in order to reach a point where the worst case
generalization error is tolerable. 

This analysis also suggests a method for evaluating the network. Let $N$ be a given DNN, and denote by $D_{N}$ the IB distortion of the network's output layer, i.e. $I(X;Y|\hat{Y})$, and by $R_{N}$ the representational complexity of the output layer, i.e. $I(X;\hat{Y})$. We can now define two measures for the performance of the network in terms of prediction and compression. The first one is the \textit{generalization gap}, 
\[
\Delta G = D_{N}-D_{IB}^{*}\left(n\right)
\]
which bounds the amount of information about $Y$ that the network
did not capture although it could have. The second measure is the
\textit{complexity gap},
\[
\Delta C = R_{N}-R^{*}\left(n\right)
\]
which bounds the amount of unnecessary complexity in the network.
Clearly, there is no reason to believe that current training algorithms
for DNNs will reach the optimal point of the IB finite sample bound.
However, we do believe that the improved feature detection along the
network's layers corresponds to improvement on the information plane
in this direction. In other words, when placing the layers of a trained
DNN on the information plane, they should form a path similar to the
green curve in figure \ref{fig-info-curve}. It is thus desirable to find new training algorithms that 
are based on the IB optimality conditions and can shift the DNN layers closer to the optimal limit.

\section{IB Phase Transitions and the Breakdown of Linear Separability}

The most intriguing aspect of our IB analysis of DNNs, which we can only begin to address here,
is its connection to the network's architecture, namely, the emergence of the layered structure
and the optimal connectivity between the layers. 

There seems to be an interesting correspondence between the IB phase transitions - 
the bifurcations to simpler representations along the information curve - and the linear 
separability condition between the hidden layers. 
Following the bifurcation analysis of the cluster splits in \cite{KR-EG-GF-1990, KR-98} for the 
IB phase transitions, one can show that the critical $\beta$ is determined by the largest eigenvalue
of the second order correlations of $p(X,Y|\hat{X}(\beta))$, at that critical $\beta$.  

\ignore{
The beginning of the curve, at the critical $\beta_0$ the representation $\hat{x}$ contains some non-zero 
information on $Y$. This means that the IB cluster centroid, \ref{IB_cent} has more than one solution.
We can identify the critical $\beta$ by a simple first order bifurcation analysis of the centroid equation.

Define $\eta\left(x,y\right)=\ln\frac{p\left(y|x\right)}{p\left(y\right)}$
and $\eta\left(\hat{x},y\right)=\ln\frac{p\left(y|\hat{x}\right)}{p\left(y\right)}$

Denoting $w\left(x,\beta\right)=\frac{p\left(x\right)e^{-\beta\left\langle \eta\left(x\right)\right\rangle _{y|x}}}{z\left(x;\beta\right)}$,
and do a Taylor expansion of \ref{IB_cent} to first order in $\eta\left(\hat{x},y\right)$, we obtain,

\begin{eqnarray*}
\eta\left(\hat{x},y\right) & \approx & \beta\sum_{y'}
\overbrace{\sum_{x}\frac{p\left(y|x\right)w\left(x,\beta\right)}{p\left(y\right)}
p\left(y'|x\right)}^{C_{yy'}}\eta\left(\hat{x},y'\right)
\end{eqnarray*}
In matrix notation
\[
0=\left(I-\beta C_{yy'}\right)\eta\left(\hat{x}\right)
\]

The last equation means that the there will be a non-zero solution
for $\eta\left(\hat{x}\right)$ when the matrix $I-\beta C_{yy'}$
has a null space. This will first happen when $\beta=\frac{1}{\lambda_{max}}$,
where $\lambda_{max}$ is the largest eigenvalue of $C_{yy'}$. $C_{yy'}$
has the form of a weighted correlation matrix of the data points $p\left(y|x\right)$.
}
On the other hand, the linear separability condition, Eq.\ref{eq:Lin-Sep}, breaks down when the  
conditional second order correlations of the data can not be ignored. This happens at the values of 
$\beta$ for which the second order (first non-linear term) of the log-likelihood ratio, conditioned on the 
current representation, $\hat{X}(\beta)$, become important, with the same eigenvalues that 
determine the phase transitions. Namely, the linear separability 
required for the DNN layers is intimately related to the structural representation phase transitions 
along the IB curve. We therefore conjecture that the optimal points for the DNN layers are 
at values of $\beta$ right after the bifurcation transitions on the IB optimal curve. When these phase transitions
are linearly independent they may be combined within a single layer, as can be done with 
linear networks (e.g. in the Gaussian IB problem \cite{GC-AG-NT-YW-2005}).  

\section{Discussion}
\ignore{
\begin{itemize}
\item we still need to check our conjecture / prediction on actual networks
\item optimal hierarchical codes  and successive refinement?
\item apply similar analysis for unsupervised learning 
\item implications on stochastic networks and especially DBNs
\end{itemize}
}
We suggest a novel information theoretic analysis of deep neural networks based on the 
information bottleneck principle. Arguably, DNNs learn to extract efficient representations of the relevant 
features of the input layer $X$ for predicting the output label $Y$, given a finite sample of the 
joint distribution $p(X,Y)$. This representation can be compered with the theoretically optimal 
relevant compression of the variable $X$ with respect to $Y$, provided
by the information bottleneck (or information distortion) tradeoff. This is done by introducing a 
new information theoretic view of DNN training as an successive (Markovian) relevant compression 
of the input variable $X$, given the empirical training data. The DNN's prediction is activating the 
trained compression layered hierarchy to generate a predicted label $\hat{Y}$. 
Maximizing the mutual information $I(Y;\hat{Y})$, for a sequence of evoking inputs $X$, 
emerges as the natural DNN optimization goal.

This new representation of DNNs offers several interesting advantages:
\begin{itemize}
\item The network and all its hidden layers can be directly compered to the optimal IB limit, 
by estimating the mutual information between each layer and the input and the output variables, on the information plane.
\item New information theoretic optimization criteria for optimal DNN representations.
\item New sample complexity bounds on the network generalization ability using the IB finite sample bounds. 
\item Stochastic DNN architectures can get closer to the optimal theoretical limit.
\item There appears to be a connection, which should be further explored, between the network architecture - the number and structure of the layers - and the structural phase transitions in the IB problem, as both are related to spectral properties of the second order correlations of the data, at the critical points. 
\end{itemize}

\addtolength{\textheight}{-12cm}




\bibliographystyle{IEEEtran}
\bibliography{IEEEabrv,deepIB_bibliography}

\end{document}